\pdfoutput=1

\documentclass[11pt]{article}

\usepackage{authblk}  

\usepackage{emnlp2021}

\newif\ifanonymous
\anonymousfalse

\usepackage{times}
\usepackage{latexsym}

\usepackage[T1]{fontenc}

\usepackage[utf8]{inputenc}

\usepackage{microtype}

\usepackage{tabularx}
\newcolumntype{L}[1]{>{\raggedright\arraybackslash}p{#1}}%
\newcolumntype{R}[1]{>{\raggedleft\arraybackslash}p{#1}}%
\newcolumntype{C}[1]{>{\centering\arraybackslash}p{#1}}%

\usepackage{subcaption}
\usepackage{float}
\usepackage{booktabs}
\usepackage[flushleft]{threeparttable}
\usepackage{multirow}
\usepackage{amsmath}
\usepackage{amssymb}
\usepackage[inline]{enumitem}
\setlist{nosep}  
\setlist[description]{leftmargin=2ex}  
\setlist[enumerate]{labelindent=0ex,labelwidth=1.5ex,leftmargin=!}
\setlist[itemize]{labelindent=0ex,labelwidth=1.5ex,leftmargin=!}
\usepackage{colortbl}
\usepackage{mathtools}

\usepackage[linesnumbered,ruled]{algorithm2e}
\SetAlCapSty{}
\SetAlCapNameFnt{\fontsize{10pt}{10pt}}
\SetAlCapFnt{\fontsize{10pt}{10pt}}
\makeatletter
\newenvironment{Ualgorithm}[1][htpb]{\def\@algocf@post@ruled{\kern\interspacealgoruled\hrule  height\algoheightrule\kern3pt\relax}%
    \def\@algocf@capt@ruled{under}%
    \setlength\algotitleheightrule{0pt}%
    \SetAlgoCaptionLayout{centerline}%
    \begin{algorithm}[#1]}
    {\end{algorithm}}
\makeatother

\usepackage{cleveref}
\crefname{section}{Section}{Sections}
\Crefname{section}{Section}{Sections}
\crefname{appendix}{Appendix}{Appendices}
\crefname{Appendix}{Appendix}{Appendices}
\crefname{subsection}{Section}{Sections}
\Crefname{subsection}{Section}{Sections}
\crefname{equation}{Equation}{Equations}
\Crefname{equation}{Equation}{Equations}
\Crefname{figure}{Figure}{Figures}
\crefname{figure}{Figure}{Figures}
\Crefname{table}{Table}{Tables}
\crefname{table}{Table}{Tables}
\crefname{enumi}{}{}
\crefname{algorithm}{Algorithm}{Algorithms}
\Crefname{algorithm}{Algorithm}{Algorithms}

\usepackage{autonum} 

\usepackage{fancyvrb}

\title{ContractNLI: A Dataset for Document-level Natural Language Inference for Contracts}

\author[12]{Yuta Koreeda}
\author[2]{Christopher D. Manning}
\affil[1]{Hitachi America Ltd, Santa Clara, CA, USA}
\affil[2]{Stanford University, Stanford, CA, USA}
\affil[ ]{{\tt \{koreeda, manning\}@stanford.edu}}
\date{}

\begin{document}
\maketitle
\begin{abstract}
    Reviewing contracts is a time-consuming procedure that incurs large expenses to companies and social inequality to those who cannot afford it.
In this work, we propose \emph{document-level natural language inference (NLI) for contracts}, a novel, real-world application of NLI that addresses such problems.
In this task, a system is given a set of hypotheses (such as ``Some obligations of Agreement may survive termination.'') and a contract, and it is asked to classify whether each hypothesis is \emph{entailed by}, \emph{contradicting to} or \emph{not mentioned by} (neutral to) the contract as well as identifying \emph{evidence} for the decision as spans in the contract.
We annotated and release the largest corpus to date consisting of 607 annotated contracts.
We then show that existing models fail badly on our task and introduce a strong baseline, which \begin{enumerate*}[label=(\arabic*)]
    \item models evidence identification as multi-label classification over spans instead of trying to predict start and end tokens, and
    \item employs more sophisticated context segmentation for dealing with long documents
\end{enumerate*}.
We also show that linguistic characteristics of contracts, such as negations by exceptions, are contributing to the difficulty of this task and that there is much room for improvement.

\end{abstract}

\section{Introduction}\label{sec:introduction}

\begin{figure}[t!]
    \centering
    \includegraphics[width=\linewidth]{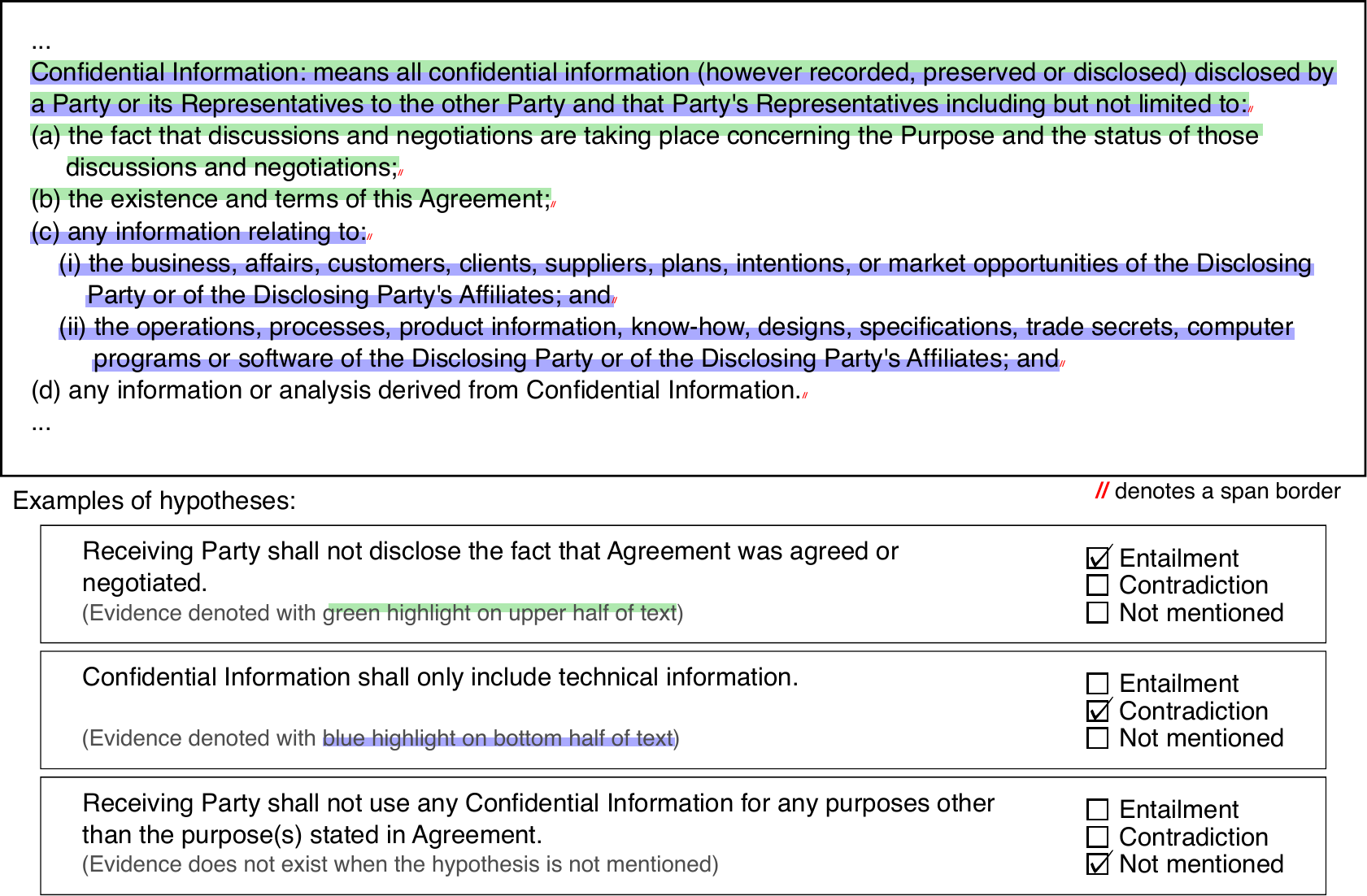}
    \caption{An overview of document-level NLI for contracts. Given a contract, a system must classify whether each hypothesis is \emph{entailed by}, \emph{contradicting to} or \emph{not mentioned by} the contract and identify \emph{evidence} for the decision as spans in the contract.}\label{fig:task_overview}
\end{figure}

Reviewing a contract is a time-consuming procedure.
A study \cite{exigent_group_limited_how_2019} revealed that ``60-80\% of all business-to-business transactions are governed by some form of written agreement, with a typical Fortune 1000 company maintaining 20,000 to 40,000 active contracts at any given time''.
Contract review is carried out manually by professionals, costing companies a huge amount of money each year.
Even worse, smaller companies or individuals may opt for signing contracts without access to such professional services.

To address this need, there is a growing interest in \emph{contract review automation}.
Recently, \citet{leivaditi_benchmark_2020} and \citet{hendrycks_cuad_2021} introduced datasets for extracting certain terms in contracts, which can help a user comprehend a contract by providing a consistent legend for what sort of terms are discussed in the contract.
However, these works only aim to find what sort of terms are present, not what each of such terms exactly states.
For example, \cite{hendrycks_cuad_2021} involves extracting a span in a contract that discusses about a question ``Is there a restriction on a party’s soliciting or hiring employees ...?''.
Being able to answer such questions can further benefit users by automatically detecting terms that are against the user's policy without having have to read each of the extracted terms.

In this paper, we argue that contract review is also a compelling real-world use case for natural language inference (NLI).
However, rather than evaluating a hypothesis versus a short passage, evaluation is against a whole document.
Concretely, given a contract and a set of hypotheses (such as ``Some obligations of Agreement may survive termination.''), we would like to classify whether each hypothesis is \emph{entailed by}, \emph{contradicting to} or \emph{not mentioned by} (neutral to) the contract as well as identifying \emph{evidence} for the decision as spans in the contract (\cref{fig:task_overview}).
Therefore, the problem involves similar evidence identification problems as open domain question answering, a problem less studied in the NLI context, and practical usefulness also involves identifying the evidence spans justifying an NLI judgment.

Our work presents a novel, real-world application of NLI.
We further argue that contracts --- which occupy a substantial amount of the text we produce today --- exhibit interesting linguistic characteristics that are worth exploring.
Our contributions are as follows:
\begin{enumerate}
    \item We annotated and release\footnote{\url{https://stanfordnlp.github.io/contract-nli/}} a dataset consisting of 607 contracts. This is the first dataset to utilize NLI for contracts and is also the largest corpus of annotated contracts.
    \item We introduce a strong baseline for our task, Span NLI BERT, which \begin{enumerate*}
        \item makes the problem of evidence identification easier by modeling the problem as multi-label classification over spans instead of trying to predict the start and end tokens, and
        \item introduces more sophisticated context segmentation to deal with long documents
    \end{enumerate*}. We show that Span NLI BERT significantly outperforms the existing models.
    \item We investigate interesting linguistic characteristics in contracts that make this task challenging even for Span NLI BERT.
\end{enumerate}

\section{ContractNLI Dataset}\label{sec:dataset}

\subsection{Task Formulation}\label{sec:system-formulation}

Our task is, given a contract and a set of hypotheses (each being a sentence), to classify whether each hypothesis is \emph{entailed by}, \emph{contradicting to} or \emph{not mentioned by} (neutral to) the contract, and to identify \emph{evidence} for the decision as spans in the contract.
More formally, the task consists of:
\begin{description}
    \item[Natural language inference (NLI)] Document-level three-class classification (one of \textsc{Entailment}, \textsc{Contradiction} or \textsc{NotMentioned}).
    \item[Evidence identification] Multi-label binary classification over \emph{span}s, where a \emph{span} is a sentence or a list item within a sentence. This is only defined when NLI label is either \textsc{Entailment} or \textsc{Contradiction}.
\end{description}
We argue that extracting whole sentences is more appropriate for ContractNLI because a lawyer can then check the evidence with comprehensible context around it, as oppose to the token-level span identification as in factoid question answering where users do not need to see the textual support for the answer.
Evidence spans therefore must be as concise as possible (need not be contiguous) while being self-contained, such that a reasonable user should be able to understand meaning just by reading the evidence spans (e.g., the second hypothesis in \cref{fig:task_overview} includes the first paragraph in order to clarify the clauses' subject).
We comprehensively identify evidence spans where they are redundant.

Unlike \cite{hendrycks_cuad_2021}, we target a single type of contracts.
This allows us to incorporate less frequent and more fine-grained hypotheses, as we can obtain a larger amount of such examples with the same number of annotated contracts.
While practioners will have to create a similar dataset to scale their system to another type of contracts, our work can be a model for how to generalize to other types of contracts because they would exhibit similar linguistic characteristics.
We chose non-disclosure agreements (NDAs) for our task, which are relatively easy to collect.

Because a lawyer would look for the same type of information in contracts of the same type, we fixed the hypotheses throughout all the contracts including the test dataset.
Given the closed set of hypotheses, this problem could also be addressed by building a text classifier for each hypothesis.
However, given the modest available data for a task requiring natural language understanding, we believe more power can be achieved by viewing this as an NLI problem.
Indeed, you can think of the NLI approach as building a multi-task text classifier with the hypothesis serving as a ``prompt'' to the model.
We will discuss whether introducing hypotheses is helpful to the model or not in \cref{sec:discussion-additional_experiments}.

\subsection{Data Collection}\label{sec:system-data}

In this section, we briefly discuss how we collected and annotated the dataset.
Since it posed many challenges that we cannot adequately describe within the page limit, we provide more details and caveats in \cref{sec:appendix-data-collection}.

\begin{figure*}[t]
    \centering
    \includegraphics{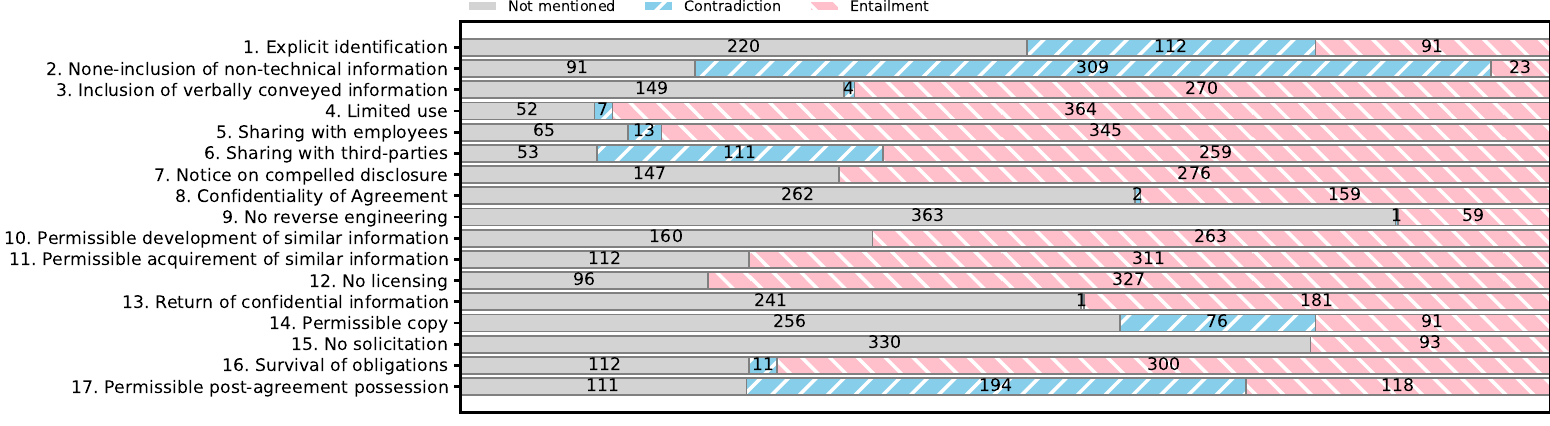}
    \caption{The NLI label distribution. Hypothesis names are used only for a human readability purpose. A full list of hypotheses can be found in \cref{tab:hypotheses}.}\label{fig:data-label_distribution}
\end{figure*}

\begin{table}[t]
    \centering
    \fontsize{8pt}{10pt}\selectfont
    \setlength{\tabcolsep}{4pt}
    \renewcommand{\arraystretch}{.6}
    \begin{tabular}{cccccc}\toprule
        Format & Source & Train & Development & Test & Total \\\midrule
        Plain Text & EDGAR & 83 & 12 & 24 & 119 \\
        HTML & EDGAR & 79 & 11 & 23 & 113\\
        PDF & Search engines & 261 & 38 & 76 & 375 \\\midrule
        \multicolumn{2}{c}{Total} & 423 & 61 & 123 & 607\\\bottomrule
    \end{tabular}
    \caption{Data split}\label{tab:data-split}
\end{table}

\begin{table}[t]
    \centering
    \fontsize{8pt}{10pt}\selectfont
    \setlength{\tabcolsep}{4pt}
    \renewcommand{\arraystretch}{.6}
    \begin{tabular}{ccccccc}\toprule
        & \multicolumn{3}{c}{Number per a document} & \multicolumn{3}{c}{Tokens per an instance} \\\cmidrule(lr){2-4}\cmidrule(lr){5-7}
        & Average & Min. & Max. & Average & Min. & Max.\\\midrule
        Paragraph & 43.7 & 9 & 248 & 52.8 & 1 & 1209 \\
        Span & 77.8 & 18 & 354 & 29.5 & 1 & 289 \\
        Token & 2,254.3 & 336 & 11,503 & --- & --- & --- \\\bottomrule
    \end{tabular}
    \caption{Basic statistics of the training dataset}\label{tab:data-tokens}
\end{table}

\begin{figure}[t]
    \centering
    \begin{subfigure}[b]{0.5\linewidth}
        \centering
        \includegraphics{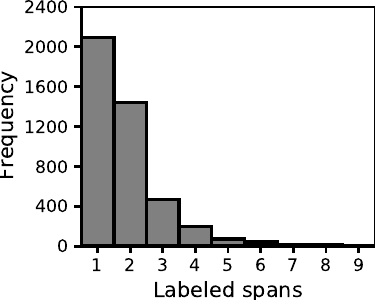}
        \caption{As number of spans}
        \label{fig:data-spans-num}
    \end{subfigure}~
    \begin{subfigure}[b]{0.5\linewidth}
        \centering
        \includegraphics{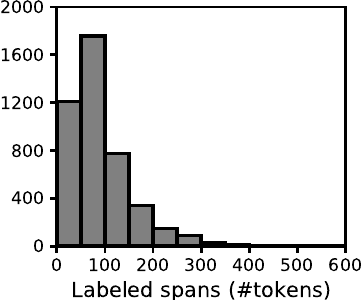}
        \caption{As number of tokens}
        \label{fig:data-spans-len}
    \end{subfigure}
    \caption{Number of evidence spans in each document.}\label{fig:data-spans}
\end{figure}

We collected NDAs from Internet search engines and Electronic Data Gathering, Analysis, and Retrieval system (EDGAR).
We searched data with a simple regular expression and hand-picked valid contracts.

Since the collected documents came in various formats including PDFs, we used \cite{koreeda_capturing_2021} to extract plain text from the documents by removing line breaks, detecting paragraph boundaries and removing headers/footers.
In order to further ensure the quality of our data, we manually screened all the documents and corrected mistakes made by the tool.
We then used Stanza \cite{qi-etal-2020-stanza} to split each paragraph into sentences and further split each sentence at inline list items (e.g., at ``(a)'' or ``iv)'') using another regular expression.
Finally, we tokenized each sentence with Stanza and further split each token into subtokens using BERT's tokenizer \cite{devlin_bert_2019,wu_googles_2016}.

For hypotheses, we developed 17 hypotheses by comparing different NDAs.
We did not include hypotheses that would simply reason about presence of certain clauses (such as ``There exists an arbitration clause in the contract.'') because they are covered by previous studies \cite{leivaditi_benchmark_2020,hendrycks_cuad_2021}.

Finally, we annotated all the contracts based on the principles discussed in \cref{sec:system-formulation}.
Since we employ a fixed set of hypotheses unlike existing NLI datasets, we were able to utilize an example-oriented annotation guideline to improve annotation consistency.

\subsection{Data Statistics}\label{sec:dataset-statistics}

We annotated a total of 607 documents, which are split into training, development and testing data at a ratio of 70:10:20 stratified by their formats (\cref{tab:data-split}).
We show statistics of the documents in \cref{tab:data-tokens}.
A document on average has 77.8 spans to choose evidence spans from.
An average number of tokens per a document is 2,254.0, which is larger than maximum allowed context length of BERT (512 tokens).
Even though an NDA is relatively short for a contract, 86\% of documents exceed the maximum allowed context length of BERT.

The distribution of NLI labels is shown in \cref{fig:data-label_distribution}.
\textsc{Entailment} and \textsc{NotMentioned} occupy a significant ratio of the dataset, but around half of the hypotheses contain both \textsc{Entailment} and \textsc{Contradiction}.
The distribution of evidence spans is shown in \cref{fig:data-spans}.
The most of entailed/contradicting hypotheses have one or two evidence spans, but some have up to nine spans.

\section{Span NLI BERT for ContractNLI}\label{sec:system-model}

\begin{figure*}[tb]
    \centering
    \includegraphics[width=\textwidth]{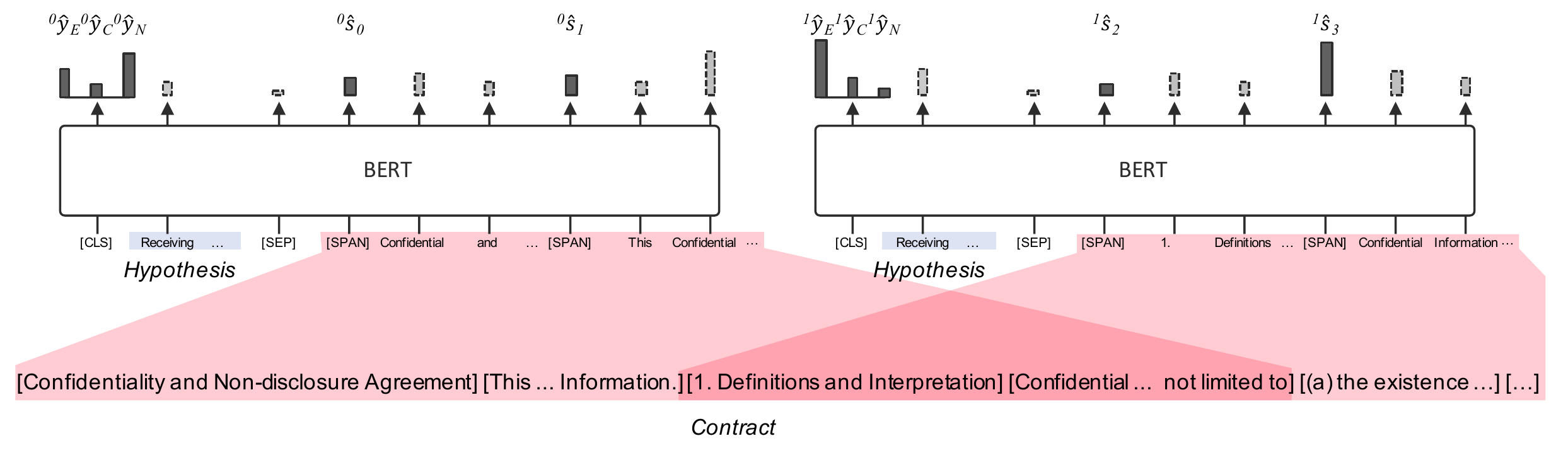}
    \caption{Model architecture of proposed Span NLI BERT}\label{fig:model}
\end{figure*}

Transformer-based models have become a dominant approach for many NLP tasks.
Previous works implemented span identification on the Transformer architecture by predicting start and end tokens, scaling it to a document by splitting the document into multiple contexts with a static window and a stride size \cite{devlin_bert_2019,hendrycks_cuad_2021}.
The start/end token prediction makes the problem unnecessarily difficult because the model has to solve span boundary detection and evidence identification concurrently, whereas the definition of spans is usually fixed for many applications.
Splitting a document can be problematic when a span is split into multiple contexts or when a span does not receive enough surrounding context.

To that end, we introduce Span NLI BERT, a multi-task Transformer model that can jointly solve NLI and evidence identification, as well as addressing the above shortcomings of the previous works (\cref{fig:model}).
Instead of predicting start and end tokens, we propose to insert special \texttt{[SPAN]} tokens each of which represents a span consisting of subsequent tokens, and model the problem as straightforward multi-label binary classification over the \texttt{[SPAN]} tokens.
We also propose to split documents with dynamic stride sizes such that there exists at least one context setting for each span in which the span is not split and receives enough context.

\begin{Ualgorithm}[t]
    \fontsize{8pt}{10pt}\selectfont
    \SetAlgoLined
    \KwIn{Span boundary token indices $B = [b_0, b_1, ...]$, Tokens $T = [t_0, t_1, ...]$, min. \# of surrounding tokens $n$, max. context length $l$}
    \KwOut{List of overlapping contexts}
    $\text{contexts} = []$ \;
    $\text{start} = 0$ \;
    \While{$\text{len}(B)$ > 0}{
        \For{$b_i$ in $B$ where $b_i - \text{start} <= l$}{
            $B$.remove($b_{i-1}$) \;
            $\text{end} = b_{i-1}$ \;
        }
        contexts.append($T[\text{start}:(\text{start} + l)]$) \;
        $\text{start} = \text{end} - \text{n}$ \;
    }
    \Return contexts \;
    \caption{Dynamic context segmentation}\label{alg:dynamic_context_segmentation}
\end{Ualgorithm}

First, we split each document into contexts using \cref{alg:dynamic_context_segmentation}.
Given a user-specified maximum context length $l$ and a minimum number of surrounding tokens $n$, the algorithm adds first $l$ tokens to a context and marks the spans whose tokens have all been added to the context.
For the next context, it will start again from $n$ tokens before the next unmarked span and repeat this until all the spans are marked.
We mark variables associated with $m$-th context with a left superscript $m$ where necessary.

For each context, contract tokens and hypothesis tokens are concatenated with a \texttt{[SEP]} token and fed into a Transformer-based model.
For evidence identification, we place a randomly initialized multi-layer perceptron (MLP) on top of each \texttt{[SPAN]} token followed by sigmoid activation to predict a span probability $\hat{s}_i  \in \mathbb{R}$.
Likewise for NLI, we place a randomly initialized MLP on top of the \texttt{[CLS]} token followed by a softmax layer to predict \textsc{Entailment},  \textsc{Contradiction} and \textsc{NotMentioned} probabilities ${}^m\hat{y}_E, {}^m\hat{y}_C, {}^m\hat{y}_N \in \mathbb{R}$, respectively.

For evidence identification loss $\ell_{span}$ of a single context, we employ cross entropy loss between the predicted span probability $\hat{s}_i$ and the ground truth span label $s_i \in \{0, 1\}$.
\begin{equation}
    \ell_{span} = \sum_i \left(- s_i \log \hat{s}_i - (1 - s_i) \log (1 - \hat{s}_i )\right)
\end{equation}
Although there exists no evidence span when NLI label is \textsc{NotMentioned}, we nevertheless incorporate such an example in the evidence identification loss with negative span labels $s_i = 0$.

For NLI loss $\ell_{NLI}$, we likewise employ cross entropy loss between the predicted NLI probabilities $\hat{y}_E, \hat{y}_C, \hat{y}_N$ and the ground truth span labels $y_E, y_C, y_N \in \{0, 1\}$.
However, there are contexts without an evidence span despite the NLI label being \textsc{Entailment} or \textsc{Contradiction}.
This causes inconsistency between what the model sees and its teacher signal.
Thus, we ignore the NLI predictions for the contexts that do not contain an evidence span.
\begin{equation}
    \ell_{NLI} = \begin{cases}
        - \sum_{L\in\{E, C ,N\}} y_L \log \hat{y}_L, & \text{if } {}^\exists s_i = 1, \\
        0, & \text{otherwise}.
    \end{cases}
\end{equation}
The multitask loss $\ell$ for a single context is then
\begin{equation}
    \ell = \ell_{span} + \lambda \ell_{NLI},
\end{equation}
where $\lambda$ is a hyperparameter that controls the balance between the two losses.
We mix contexts from different documents during training, thus contexts from a single document may appear in different mini batches.

Since each document is predicted as multiple contexts, results from these contexts have to be aggregated to obtain a single output for a document.
For the evidence identification, we simply take the average of span probabilities over different model outputs.
\begin{equation}
    {}^\ast\hat{s}_i = \frac{1}{M_i} \sum_m {}^m\hat{s}_i,
\end{equation}
where $M_i$ is the number of contexts that have the full $i$-th span in its context.

For NLI, we weighted the NLI probabilities by the sum of the span probabilities:
\begin{equation}
    {}^\ast\hat{y}_\bullet =  \frac{1}{\sum_m \frac{1}{S_m} \sum_i {}^m\hat{s}_i}\sum_m \left({}^m\hat{y}_\bullet \cdot \frac{1}{S_m} \sum_i {}^m\hat{s}_i\right),
\end{equation}
where $S_m$ is the number of \texttt{[SPAN]} tokens in the $m$-th context.
This is based on an intuition that contexts with evidence spans should contribute more to NLI.

\section{Experiments}\label{sec:experiment}

\subsection{Baselines}\label{sec:baselines}

In order to study the dataset's characteristics, we implemented five baselines with different capabilities.
We briefly explain the five baselines that we implemented below, but more details can be found in \cref{sec:appendix-experiments-baseline}

\begin{description}
    \item[Majority vote] A baseline that outputs an oracle majority label for each hypothesis (NLI only).
    \item[Doc TF-IDF+SVM] A document-level multi-class linear Support Vector Machine \cite[SVM;][]{chang_libsvm_2011} with unigram bag-of-words features (NLI only).
    \item[Span TF-IDF+Cosine] Evidence identification based on unigram TF-IDF cosine similarities between each hypothesis and each span  (evidence identification only).
    \item[Span TF-IDF+SVM] A span-level binary Linear SVM with unigram bag-of-words features  (evidence identification only).
    \item[SQuAD BERT] A Transformer-based model as in the previous works discussed in \cref{sec:system-model}.
    Instead of allowing it to predict spans at arbitrary boundaries, we calculate a score for each of predefined spans by averaging token scores associated with the start and end of the span over different context windows.
    This makes sure that its performance is not discounted for getting span boundaries wrong.
\end{description}

\subsection{Experiment Settings}\label{sec:experiment-settings}

For evidence identification, we report mean average precision (mAP) that is micro averaged over labels.
We also report precision at recall 0.8 (P@R80) that is micro averaged over documents and labels.
P@R80 is the precision score when the threshold for evidence identification is adjusted to achieve a recall score of 0.8.
It was used in \cite{hendrycks_cuad_2021} to measure efficacy of a system under a required coverage level that is similar to typical human's.

\begin{table*}[t]
    \centering
    \begin{threeparttable}
        \centering
        \fontsize{8pt}{10pt}\selectfont
        \setlength{\tabcolsep}{4pt}
        \renewcommand{\arraystretch}{.6}
        \begin{tabular}{llc@{\hskip0pt}cc@{\hskip0pt}cc@{\hskip0pt}cc@{\hskip0pt}cc@{\hskip0pt}c}\toprule
            &  & \multicolumn{4}{c}{Evidence} & \multicolumn{6}{c}{NLI} \\\cmidrule(lr){3-6}\cmidrule(lr){7-12}
            Backbone Model & Fine-tuning Method & \multicolumn{2}{c}{mAP} & \multicolumn{2}{c}{P@R80} & \multicolumn{2}{c}{Acc.} & \multicolumn{2}{c}{F1 (C)} & \multicolumn{2}{c}{F1 (E)} \\\midrule
            BERT\textsubscript{\textit{base}} & None & .885 & \textsubscript{.025} & .663 & \textsubscript{.093} & .838 & \textsubscript{.020} & .287 & \textsubscript{.022} & .765 & \textsubscript{.035} \\
            BERT\textsubscript{\textit{large}} & None & .922 & \textsubscript{.006} & .793 & \textsubscript{.018} & .875 & \textsubscript{.006} & .357 & \textsubscript{.039} & .834 & \textsubscript{.002} \\
            DeBERTa v2\textsubscript{\textit{xlarge}} & None & .933 & \textsubscript{.002} & .859 & \textsubscript{.008} & .885 & \textsubscript{.001} & .360 & \textsubscript{.027} & .855 & \textsubscript{.002} \\
            \midrule
            BERT\textsubscript{\textit{base}} & Pretrained from scratch using a case law corpus \cite{zheng_when_2021} & .870 & \textsubscript{.015} & .578 & \textsubscript{.052} & .831 & \textsubscript{.032} & .289 & \textsubscript{.026} & .783 & \textsubscript{.040} \\
            BERT\textsubscript{\textit{base}} & Fine-tuned on case law and contract corpora \cite{chalkidis_legal-bert_2020} & .925 & \textsubscript{.004} & .811 & \textsubscript{.002} & .794 & \textsubscript{.008} & .272 & \textsubscript{.008} & .746 & \textsubscript{.018} \\
            DeBERTa v2\textsubscript{\textit{xlarge}} & Fine-tuned on span identification \cite{hendrycks_cuad_2021} & .936 & \textsubscript{.002} & .860 & \textsubscript{.003} & .892 & \textsubscript{.001} & .405 & \textsubscript{.016} & .859 & \textsubscript{.005} \\
            \midrule
            BERT\textsubscript{\textit{base}} & Fine-tuned on NDAs & .892 & \textsubscript{.002} & .690 & \textsubscript{.014} & .864 & \textsubscript{.004} & .326 & \textsubscript{.014} & .820 & \textsubscript{.010} \\
            BERT\textsubscript{\textit{large}} & Fine-tuned on NDAs & .922 & \textsubscript{.003} & .837 & \textsubscript{.008} & .875 & \textsubscript{.000} & .389 & \textsubscript{.009} & .839 & \textsubscript{.003} \\
            \bottomrule
        \end{tabular}
        \makeatletter\def\TPT@hsize{}\makeatletter
        \begin{tablenotes}[para,flushleft]
            \raggedright
            \fontsize{7pt}{7pt}\selectfont
            Refer to \cref{sec:experiment-settings} for the details on the metrics.
        \end{tablenotes}
    \end{threeparttable}
    \caption{Results for different backbone and pretrained models}\label{tab:results-pretrained}
\end{table*}

\begin{table}[t]
    \centering
    \begin{threeparttable}
        \centering
        \fontsize{8pt}{10pt}\selectfont
        \setlength{\tabcolsep}{2pt}
        \renewcommand{\arraystretch}{.6}
        \begin{tabular}{lc@{\hskip0pt}cc@{\hskip0pt}cc@{\hskip0pt}cc@{\hskip0pt}cc@{\hskip0pt}c}\toprule
            & \multicolumn{4}{c}{Evidence} & \multicolumn{6}{c}{NLI} \\\cmidrule(lr){2-5}\cmidrule(lr){6-11}
            & \multicolumn{2}{c}{mAP} & \multicolumn{2}{c}{P@R80} & \multicolumn{2}{c}{Acc.} & \multicolumn{2}{c}{F1 (C)} & \multicolumn{2}{c}{F1 (E)} \\\midrule
            Majority vote & \multicolumn{2}{c}{---} & \multicolumn{2}{c}{---} & .674 &  & .083 &  & .428 & \\
            Doc TF-IDF+SVM & \multicolumn{2}{c}{---} & \multicolumn{2}{c}{---} & .733 &  & .197 &  & .641 & \\
            Random & .024 & & .000 &  & \multicolumn{2}{c}{---} & \multicolumn{2}{c}{---} & \multicolumn{2}{c}{---} \\
            Span TF-IDF+Cosine & .381 & & .057  &  & \multicolumn{2}{c}{---} & \multicolumn{2}{c}{---} & \multicolumn{2}{c}{---} \\
            Span TF-IDF+SVM & .836 & & .322 &  & \multicolumn{2}{c}{---} & \multicolumn{2}{c}{---} & \multicolumn{2}{c}{---} \\
            SQuAD (BERT\textsubscript{\textit{base}}) & .825 & \textsubscript{.004} & .574 & \textsubscript{.004} & \multicolumn{2}{c}{---} & \multicolumn{2}{c}{---} & \multicolumn{2}{c}{---} \\
            SQuAD (BERT\textsubscript{\textit{large}}) & .869 & \textsubscript{.005} & .661 & \textsubscript{.043} & \multicolumn{2}{c}{---} & \multicolumn{2}{c}{---} & \multicolumn{2}{c}{---} \\
            \midrule
            Ours (BERT\textsubscript{\textit{base}})  & .885 & \textsubscript{.025} & .663 & \textsubscript{.093} & .838 & \textsubscript{.020} & .287 & \textsubscript{.022} & .765 & \textsubscript{.035} \\
            Ours (BERT\textsubscript{\textit{large}}) & .922 & \textsubscript{.006} & .793 & \textsubscript{.018} & .875 & \textsubscript{.006} & .357 & \textsubscript{.039} & .834 & \textsubscript{.002} \\
            \bottomrule
        \end{tabular}
        \makeatletter\def\TPT@hsize{}\makeatletter
        \begin{tablenotes}[para,flushleft]
            \raggedright
            \fontsize{7pt}{7pt}\selectfont
            Refer to \cref{sec:experiment-settings} for the details on the metrics.
        \end{tablenotes}
    \end{threeparttable}
    \caption{Main results}\label{tab:results}
\end{table}

For NLI, we report accuracy, a F1 score for contradiction (F1 (C)) and for entailment (F1 (E)).
We micro average these scores over documents and then macro average over labels.
This is to avoid the label imbalance to \emph{cancel out} with micro averaging and the results to appear too optimistic.

For our Span NLI BERT, we ran the same experiment ten times with different hyperparameters (detailed in \cref{sec:appendix-experiments-hyperparameters}) and report the average score of three models with the best development scores.
Since NLI is more challenging than evidence identification, we used macro average NLI accuracy for the criterion.
For the SQuAD BERT baseline, we ran hyperparameter search over 18 hyperparameter sets as described in \cite{devlin_bert_2019} and likewise report the average score of the three best models.
The metrics for the experiments with the hyperparameter search are followed by subscript numbers each of which denotes standard deviation of metrics over three runs.

\subsection{Results}

We first compared Span NLI BERT against baselines (\cref{tab:results}).
Span NLI BERT performed significantly better than the baselines, both in terms of evidence identification and NLI.
Nevertheless, the performance for contradiction labels is much worse than that of entailment labels, due to the imbalanced label distribution.
In terms of evidence identification, SQuAD BERT's mAP score was no better than that of Span TF-IDF+SVM, which illustrates the importance of explicitly incorporating span boundaries to input.

We then compared Span NLI BERT's performance with different backbone models and pretraining corpora including DeBERTa v2 \cite{he_deberta_2021} which was most successful in \cite{hendrycks_cuad_2021} (\cref{tab:results-pretrained}).
We can observe that making the models bigger benefits both evidence identification and NLI.
Fine-tuning models on legal corpora had mixed results.
Using a model pretrained on a case law corpus \cite{zheng_when_2021} did not benefit evidence identification nor NLI.
Fine-tuning BERT\textsubscript{\textit{base}} on NDAs has slightly improved the performance but the benefit is no longer visible for BERT\textsubscript{\textit{large}}.
Transferring DeBERTa\textsubscript{\textit{xlarge}} trained on CUAD \cite{hendrycks_cuad_2021} gave marginal improvement on NLI, making it the best performing model on the ContractNLI dataset.

\begin{table}[t]
    \centering
    \begin{threeparttable}
        \centering
        \fontsize{8pt}{10pt}\selectfont
        \setlength{\tabcolsep}{2pt}
        \renewcommand{\arraystretch}{.6}
        \begin{tabular}{lc@{\hskip0pt}cc@{\hskip0pt}cc@{\hskip0pt}cc@{\hskip0pt}cc@{\hskip0pt}c}\toprule
            & \multicolumn{4}{c}{Evidence} & \multicolumn{6}{c}{NLI} \\\cmidrule(lr){2-5}\cmidrule(lr){6-11}
            Hypothesis Usage & \multicolumn{2}{c}{mAP} & \multicolumn{2}{c}{P@R80} & \multicolumn{2}{c}{Acc.} & \multicolumn{2}{c}{F1 (C)} & \multicolumn{2}{c}{F1 (E)} \\\midrule
            Symbol (BERT\textsubscript{\textit{base}})  & .857 & \textsubscript{.044} & .574 & \textsubscript{.136} & .830 & \textsubscript{.014} & .294 & \textsubscript{.075} & .751 & \textsubscript{.027} \\
            Symbol (BERT\textsubscript{\textit{large}}) & .894 & \textsubscript{.020} & .703 & \textsubscript{.092} & .849 & \textsubscript{.006} & .303 & \textsubscript{.058} & .794 & \textsubscript{.026} \\
            \midrule
            Text (BERT\textsubscript{\textit{base}})  & .885 & \textsubscript{.025} & .663 & \textsubscript{.093} & .838 & \textsubscript{.020} & .287 & \textsubscript{.022} & .765 & \textsubscript{.035} \\
            Text (BERT\textsubscript{\textit{large}}) & .922 & \textsubscript{.006} & .793 & \textsubscript{.018} & .875 & \textsubscript{.016} & .357 & \textsubscript{.039} & .834 & \textsubscript{.002} \\
            \bottomrule
        \end{tabular}
        \makeatletter\def\TPT@hsize{}\makeatletter
        \begin{tablenotes}[para,flushleft]
            \raggedright
            \fontsize{7pt}{7pt}\selectfont
            Refer to \cref{sec:experiment-settings} for the details on the metrics.
        \end{tablenotes}
    \end{threeparttable}
    \caption{A controlled experiment using a randomly initialized special token for each hypothesis (Symbol) instead of hypothesis' surface tokens (Text)}\label{tab:results-symbolic}
\end{table}

\section{Discussion}\label{sec:discussion}

\subsection{Controlled Experiments}\label{sec:discussion-additional_experiments}

In order to identify what is and what is not capable by the models, we carried out controlled experiments where we modified the input of the models.

\paragraph{Is Hypothesis Information Useful?}

It is non-trivial that hypotheses surface tokens which were merely used as an instruction to the annotators can be useful in evidence identification.
The fact that Span TF-IDF+Cosine performed significantly better than the random baseline (\cref{tab:results}) implies that hypothesis surface tokens do convey useful information.
Furthermore, we also experimented with a condition where we used a randomly initialized special token for each hypothesis instead of the hypothesis' surface tokens.
Removing the hypothesis surface tokens resulted in consistent decrease of performance for both evidence identification and NLI (\cref{tab:results-symbolic}).
This implies that the hypothesis surface tokens are somewhat meaningful, but these narrow differences suggest that there could be a better way to utilize the hypothesis surface tokens.

\begin{table}[t]
    \centering
    \begin{threeparttable}
        \centering
        \fontsize{8pt}{10pt}\selectfont
        \setlength{\tabcolsep}{3pt}
        \renewcommand{\arraystretch}{.6}
        \begin{tabular}{lc@{\hskip0pt}cc@{\hskip0pt}cc@{\hskip0pt}c}\toprule
            & \multicolumn{6}{c}{NLI} \\\cmidrule(lr){2-7}
            &  \multicolumn{2}{c}{Accuracy} & \multicolumn{2}{c}{F1 (C)} & \multicolumn{2}{c}{F1 (E)} \\\midrule
            Majority vote & .814 &  & .239 &  & .645 & \\
            Span NLI (BERT\textsubscript{\textit{base}}) & .883 & \textsubscript{.006} & .490 & \textsubscript{.007} & .795 & \textsubscript{.005} \\
            Span NLI (BERT\textsubscript{\textit{large}}) & .899 & \textsubscript{.004} & .492 & \textsubscript{.065} & .820 & \textsubscript{.012} \\
            Oracle NLI (BERT\textsubscript{\textit{base}}) & .918 & \textsubscript{.005} & .657 & \textsubscript{.062} & .816 & \textsubscript{.006} \\
            Oracle NLI  (BERT\textsubscript{\textit{large}}) & .908 & \textsubscript{.011} & .620 & \textsubscript{.082} & .806 & \textsubscript{.015} \\
            \bottomrule
        \end{tabular}
        \makeatletter\def\TPT@hsize{}\makeatletter
        \begin{tablenotes}[para,flushleft]
            \raggedright
            \fontsize{7pt}{7pt}\selectfont
            Refer to \cref{sec:experiment-settings} for the details on the metrics.
        \end{tablenotes}
    \end{threeparttable}
    \caption{A controlled experiment of document-level binary classification over \textsc{Entailment} and \textsc{Contradiction} utilizing oracle evidence spans.}\label{tab:results-nli_only}
\end{table}

\paragraph{Can Better Evidence Identification Lead to Better NLI?}

In ContractNLI, evidence identification and NLI are dependent on each other.
We experimented whether good evidence identification can benefit NLI by feeding models with oracle evidence spans for NLI.
For the oracle model (Oracle NLI), we concatenated a hypothesis and ground truth evidence spans as an input and predicted a binary label of \textsc{Entailment} or \textsc{Contradiction}.
We can observe in \cref{tab:results-nli_only} that giving models oracle spans substantially improves NLI performance, notably the F1 score of \textsc{Contradiction}.
This suggests that there is still much room for improvement on NLI just by improving evidence identification.

\begin{table}[t]
    \centering
    \begin{threeparttable}
        \centering
        \fontsize{8pt}{10pt}\selectfont
        \setlength{\tabcolsep}{3pt}
        \renewcommand{\arraystretch}{.6}
        \begin{tabular}{lcccc}\toprule
             & \multicolumn{3}{c}{NLI Accuracy} & \\\cmidrule{2-4}
            Condition & Majority & Minority & Weighted & \% minority label \\\midrule
            w/o (local) & .91 & .77 & .84 & 21 \\
            w/ (local) & .92 &  .40 & .66 & 7 \\
            \midrule
            w/o (non-local) & .98 & .72 & .85 & 19 \\
            w/ (non-local) & .90 & .00 & .45 & 6  \\
            \bottomrule
        \end{tabular}
        \makeatletter\def\TPT@hsize{}\makeatletter
        \begin{tablenotes}[para,flushleft]
            \raggedright
            \fontsize{7pt}{7pt}\selectfont
            Accuracy has been calculated for majority and minority ground-truth NLI labels separately in order to rule out the effect of the label distribution. ``Weighted'' denotes an average of the two accuracy scores that are weighted disproportionally to the number of occurances of each label. Only the hypotheses that exhibit negation by exception are used for this experiment (\#1, 4, 5, 6, 14, 16 and 17 for local, and \#4, 5, 6 and 17 for non-local).
        \end{tablenotes}
    \end{threeparttable}
    \caption{NLI accuracy in instances with and without (non-)local negation by exception}\label{tab:experiment-exception}
\end{table}

\subsection{Challenges of ContractNLI}\label{sec:discussion-observations}

Our task is challenging from a machine learning perspective.
The label distribution is imbalanced and it is naturally multi-task, all the while training data being scarce.
Furthermore, we argue that there exist multiple linguistic characteristics of contracts that make the task challenging.

We annotated the development dataset on whether each document-hypothesis pair exhibits certain characteristics and evaluated impact of each characteristic on the performance of the best Span NLI BERT (BERT\textsubscript{\textit{large}}) from \cref{tab:results}.
Since evidence spans are only available when the NLI label is either \textsc{Entailment} or \textsc{Contradiction}, document-hypothesis pairs with \textsc{NotMentioned} label are excluded from the evaluations in this section.

\paragraph{Negation by Exception}

Contracts often state a general condition and subsequently add exceptions to the general condition.
For example, in ``Recipient shall not disclose Confidential Information to any person or entity, except its employees or partners ...'', the first half clearly forbids sharing confidential information to an employee, but the latter part flips this decision and it is actually permitting the party to share confidential information.
This phenomenon can occur both locally (i.e., within a single span) or non-locally, sometimes pages away from each other.
In our dataset, the local case happens in 12\% of document-hypothesis pairs, which corresponds to 59\% of documents with at least one of such hypotheses.
The non-local case happens in 7\% of document-hypothesis pairs and 44\% of documents.
By comparing document-hypothesis pairs with and without such phenomena, we can see that local and non-local negation by exception is hurting the model's NLI accuracy (\cref{tab:experiment-exception}).

\begin{table}[t]
    \centering
    \begin{threeparttable}
        \centering
        \fontsize{8pt}{10pt}\selectfont
        \setlength{\tabcolsep}{3pt}
        \renewcommand{\arraystretch}{.6}
        \begin{tabular}{llcccc}\toprule
            & &  & \multicolumn{2}{c}{\# spans read before finding: } & \\\cmidrule(lr){4-5}
            & $n$ & \# spans & one span & all spans &  mAP \\\midrule
            Continuous & 128 & 2.64 & 1.09 & 3.82 & 0.91 \\
            Discontinuous & 128 & 2.34 & 1.04 & 3.84 & 0.94 \\
            \midrule
            Continuous & 64 & 2.64 & 1.16 & 4.33 & 0.89 \\
            Discontinuous & 64 & 2.34 & 1.01 & 4.85 & 0.94 \\
            \bottomrule
        \end{tabular}
        \makeatletter\def\TPT@hsize{}\makeatletter
        \begin{tablenotes}[para,flushleft]
            \raggedright
            \fontsize{7pt}{7pt}\selectfont
            ``\# spans read before finding one (all) span(s)'' refers to the number of spans a user needs to read until the user finds one (all) span(s) if the user reads the spans in an order of a system's span probability output. Thus, it is better when it is lower and 1.0 is the best possible value.
        \end{tablenotes}
    \end{threeparttable}
    \caption{Evidence identification performance of models with different minimum number of surrounding tokens $n$ on documents with dis-/continuous spans}\label{tab:experiment-discontinous}
\end{table}

\paragraph{Discontinous Spans}

As sketched in \cref{fig:task_overview}, evidence spans can be discontinous and may even be pages apart.
Such discontinous spans occur in 28\% of document-hypothesis pairs, which corresponds to 81\% of documents with at least one of such hypotheses.

Contrary to our expectation, discontinuous setting did not have a negative effect on overall evidence identification mAP score (\cref{tab:experiment-discontinous}).
This can be attributed to the fact finding a single span was easier in the discontinuous setting, which is evident from ``the number of spans read before finding one span''.
``Number of spans read before finding all spans'' is nevertheless affected by discontinous spans, especially when the model's minimum number of surrounding tokens $n$ is small\footnote{This is the best BERT\textsubscript{large} with $n=64$ and the fifth best model overall.}.
Furthermore, there was a positive correlation between the gap between the discontinuous spans and ``number of spans read before finding all spans'' (a Spearman correlation of $\rho=0.205$, $p=0.015$).
This is because many hypothesis-distinctive spans (e.g., a span starting with ``(ii)'' in the second hypothesis of \cref{fig:task_overview}) can be inferred without access to its context, but finding the accompanying spans (e.g., the first span in \cref{fig:task_overview}) is impossible when they do not fit onto a single context window.
Nevertheless, the effect of discontinous spans is very small and Span NLI BERT can overcome this with a larger number of surrounding tokens.

\paragraph{Reference to Definition}

\begin{table}[t]
    \centering
    \begin{threeparttable}
        \centering
        \fontsize{8pt}{10pt}\selectfont
        \setlength{\tabcolsep}{3pt}
        \renewcommand{\arraystretch}{.6}
        \begin{tabular}{lcccc}\toprule
            & \multicolumn{3}{c}{NLI Accuracy} & \\\cmidrule{2-4}
            Condition & Majority & Minority & Weighted & \% minority label \\\midrule
            w/o Reference & .91 & .88 & .89 & 26 \\
            w/ Reference & .93 &  --- & --- & 0 \\
            \bottomrule
        \end{tabular}
        \makeatletter\def\TPT@hsize{}\makeatletter
        \begin{tablenotes}[para,flushleft]
            \raggedright
            \fontsize{7pt}{7pt}\selectfont
            Accuracy has been calculated for majority and minority ground-truth NLI labels separately in order to rule out the effect of the label distribution. ``Weighted'' denotes an average of the two accuracy scores that are weighted disproportionally to the number of occurances of each label. Only the hypotheses that exhibit references are used in this experiment (\#5 and 6).
        \end{tablenotes}
    \end{threeparttable}
    \caption{NLI accuracy on documents with and without references to definitions}\label{tab:experiment-reference}
\end{table}

Contracts often have references to definitions.
In our dataset, hypotheses \#5 and 6 ``Sharing with employees/third-parties'' tend to have such references.
For example, if a contract says ``The Receiving Party undertakes to permit access to the Confidential Information only to its Representatives ....'', the hypothesis \#5 ``Sharing with employees'' is entailed by such span but the hypothesis \#6 ``Sharing with third-parties'' is not.
Only when the contract includes a definition such as ``\,``Representatives'' shall mean directors, employees, professional advisors or anyone involved with the Party in a professional or business capacity.'', hypothesis \#6 is also entailed by the contract.
We speculated that this could make NLI more difficult because the model has to refer to both spans in order to get NLI right.
However, our observation discovered that examples with references are no more difficult than those without them (\cref{tab:experiment-reference}).

\section{Related Works}

Helped by their accessibility, there exist multiple prior works on ``legal NLI'' for case and statute laws.
One of the subtasks in COLIEE-2020 shared task \cite{rabelo_coliee_nodate} was, given a court decision Q and relevant cases, to extract relevant paragraphs from the cases and to classify whether those paragraphs entail ``Q'' or ``not Q''.
\citet{holzenberger_dataset_2020} introduced a dataset for predicting an entailment relationship between a statement and a statute excerpt.
While they are both ``legal'' and ``NLI'', statutes and contracts exhibit different characteristics including the fact that statutes/cases tend to be written in consistent vocabulary and styles.
Moreover, there only exists a single right answer for a hypothesis in case/statute law NLI, whereas a hypothesis can be entailed by or contradicting to each contract in our task; i.e., hypotheses and documents have one-to-one relationships in case/statute law NLI, but they have many-to-many relationships in our task.

As discussed in \cref{sec:introduction}, our task has practical and scientific significance compared to information extraction for contracts \cite{leivaditi_benchmark_2020,hendrycks_cuad_2021}.
We showed in our experiments that the NLI part of our task is much more challenging than the evidence identification task.
Furthermore, we gave observations to linguistic characteristics of our dataset that are lacking in these prior works.

\citet{lippi_claudette_2019} presented a dataset where certain types of contract clauses are identified and annotated with ``clearly fair'', ``potentially unfair'' or ``clearly unfair''.
While the format of the task input and output is quite similar, our task requires reasoning over a much diverse set of hypotheses than just fair or unfair.
Similarly, fact extraction and claim verification tasks \cite{thorne_fever_2018,jiang_hover_2020}, where the task is to extract facts from Wikipedia articles and to classify whether the claim is entailed by the facts, have similar input and output formats.
Such claims and our hypotheses are quite different in nature and working on contracts poses unique challenges as discussed in \cref{sec:discussion-observations}.

\section{Conclusion}

In this work, we introduced a novel, real-world application of NLI, \emph{document-level NLI for contracts} which aim to assist contract review.
We annotated a dataset consisting of 607 contracts and showed that linguistic characteristics of contracts, particularly negations by exceptions, make the problem difficult.

We introduced Span NLI BERT that incorporates more natural solution to evidence identification by modeling the problem as multi-label classification over spans instead of trying to predict the start and the end token as in previous works.
Span NLI BERT performed significantly better than existing Transformer-based models.

Notwithstanding the performance gain by Span NLI BERT, there exists much room for improvement.
Span NLI BERT still has poor performance on rare labels, as well as being easily impacted by negations by exceptions.

For future works, we will also explore systems that can generalize to different types of contracts and hypotheses.
We believe that studying how hypothesis phrasing can affect performance and developing a better way to utilize hypothesis text can be the key to such goal.

We hope that the dataset and Span NLI BERT will serve as a starting point for tackling the interesting challenges in our ContractNLI task.

\section*{Ethical Consideration}

In this work, we collected contracts from EDGAR and Internet search engines.
For the former, EDGAR states that all filed documents are public information and can be redistributed without a further consent\footnote{\url{https://www.sec.gov/privacy.htm\#dissemination}}.
For the latter, we obtained publicly accessible documents and our academic use is within the scope of fair use.
Nevertheless, we placed a contact form for a concerned individual or organization in a similar way as other crawled datasets.

For the annotation, we hosted our annotation task on Amazon Mechanical Turk so that each worker can participate voluntarily and withdraw at any time.
We made sure each worker receives at least the US federal wage and the actual average pay was 18.31 US dollars per hour (excluding Amazon Mechanical Turk fees).
Our annotation procedure did not go through an institutional review board since we are not directly collecting information from human subjects.

While we did not run computationally expensive pretraining of Transformer-based models, we ran fine-tuning of the models 156 times for this paper.
Running experiments multiple times was necessary in order to ensure validity and reproducibility of the experiments when our dataset is modest in size from a machine learning perspective.
We believe this energy consumption can be justified by resources that we can potentially save by assisting contract review.
Moreover, we introduced an architectural change that benefits the models more than simply making the model larger (e.g., Span NLI BERT with BERT\textsubscript{\textit{base}} performed better than SQuAD BERT with BERT\textsubscript{\textit{large}} in \cref{tab:results}).

There was a concern that publication of our annotations or models may be regarded as an unauthorized practice of law (i.e., giving a legal advice without a license), which is forbidden in many jurisdictions.
This also means that an individual may suffer from a loss by relying on information from our annotation or model outputs as a legal advice.
We have consulted an attorney regarding this issue and were advised that releasing general information (the annotations and the models) does not constitute an unauthorized practice of law.
We were nevertheless advised to place a disclaimer that warns users not to rely on the information and to seek an attorney's advice instead.
Furthermore, we took additional measures, such as forbidding a crawler to index our annotations, in order to minimize a risk of an individual from referencing our annotation as a legal advice.

\ifanonymous \else \section*{Acknowledgements}

We used computational resource of AI Bridging Cloud Infrastructure (ABCI) provided by the National Institute of Advanced Industrial Science and Technology (AIST) for the experiments.
 \fi

\bibliography{main}

\begin{thebibliography}{18}
\expandafter\ifx\csname natexlab\endcsname\relax\def\natexlab#1{#1}\fi

\bibitem[{Chalkidis et~al.(2020)Chalkidis, Fergadiotis, Malakasiotis, Aletras,
  and Androutsopoulos}]{chalkidis_legal-bert_2020}
Ilias Chalkidis, Manos Fergadiotis, Prodromos Malakasiotis, Nikolaos Aletras,
  and Ion Androutsopoulos. 2020.
\newblock \href {https://doi.org/10.18653/v1/2020.findings-emnlp.261}
  {{LEGAL}-{BERT}: {The} {Muppets} straight out of {Law} {School}}.
\newblock In \emph{Findings of the {Association} for {Computational}
  {Linguistics}: {EMNLP} 2020}, pages 2898--2904.

\bibitem[{Chang and Lin(2011)}]{chang_libsvm_2011}
Chih-Chung Chang and Chih-Jen Lin. 2011.
\newblock \href {https://doi.org/10.1145/1961189.1961199} {{LIBSVM}: {A}
  {Library} for {Support} {Vector} {Machines}}.
\newblock \emph{ACM Transactions on Intelligent Systems and Technology}, 2(3).

\bibitem[{Devlin et~al.(2019)Devlin, Chang, Lee, and
  Toutanova}]{devlin_bert_2019}
Jacob Devlin, Ming-Wei Chang, Kenton Lee, and Kristina Toutanova. 2019.
\newblock \href {https://doi.org/10.18653/v1/N19-1423} {{BERT}: {Pre}-training
  of {Deep} {Bidirectional} {Transformers} for {Language} {Understanding}}.
\newblock In \emph{Proceedings of the 2019 {Conference} of the {North}
  {American} {Chapter} of the {Association} for {Computational} {Linguistics}:
  {Human} {Language} {Technologies}}.

\bibitem[{{Exigent Group Limited}(2019)}]{exigent_group_limited_how_2019}
{Exigent Group Limited}. 2019.
\newblock \href
  {https://offers.exigent-group.com/how-gcs-can-thrive-not-just-survive} {How
  {GCs} can thrive, not just survive}.
\newblock Technical report.

\bibitem[{He et~al.(2021)He, Liu, Gao, and Chen}]{he_deberta_2021}
Pengcheng He, Xiaodong Liu, Jianfeng Gao, and Weizhu Chen. 2021.
\newblock \href {http://arxiv.org/abs/2006.03654} {{DeBERTa}:
  {Decoding}-enhanced {BERT} with {Disentangled} {Attention}}.
\newblock \emph{arXiv:2006.03654 [cs]}.

\bibitem[{Hendrycks et~al.(2021)Hendrycks, Burns, Chen, and
  Ball}]{hendrycks_cuad_2021}
Dan Hendrycks, Collin Burns, Anya Chen, and Spencer Ball. 2021.
\newblock \href {http://arxiv.org/abs/2103.06268} {{CUAD}: {An}
  {Expert}-{Annotated} {NLP} {Dataset} for {Legal} {Contract} {Review}}.
\newblock \emph{arXiv}.

\bibitem[{Holzenberger et~al.(2020)Holzenberger, Blair-Stanek, and
  Van~Durme}]{holzenberger_dataset_2020}
Nils Holzenberger, Andrew Blair-Stanek, and Benjamin Van~Durme. 2020.
\newblock A {Dataset} for {Statutory} {Reasoning} in {Tax} {Law} {Entailment}
  and {Question} {Answering}.
\newblock In \emph{Proceedings of the 2020 {Natural} {Legal} {Language}
  {Processing} ({NLLP}) {Workshop}}.

\bibitem[{Jiang et~al.(2020)Jiang, Bordia, Zhong, Dognin, Singh, and
  Bansal}]{jiang_hover_2020}
Yichen Jiang, Shikha Bordia, Zheng Zhong, Charles Dognin, Maneesh Singh, and
  Mohit Bansal. 2020.
\newblock \href {https://doi.org/10.18653/v1/2020.findings-emnlp.309} {{HoVer}:
  {A} {Dataset} for {Many}-{Hop} {Fact} {Extraction} {And} {Claim}
  {Verification}}.
\newblock In \emph{Findings of the {Association} for {Computational}
  {Linguistics}: {EMNLP} 2020}, pages 3441--3460. Association for Computational
  Linguistics.

\bibitem[{Koreeda and Manning(2021)}]{koreeda_capturing_2021}
Yuta Koreeda and Christopher~D. Manning. 2021.
\newblock \href {http://arxiv.org/abs/2105.00150} {Capturing {Logical}
  {Structure} of {Visually} {Structured} {Documents} with {Multimodal}
  {Transition} {Parser}}.
\newblock \emph{arXiv}.

\bibitem[{Leivaditi et~al.(2020)Leivaditi, Rossi, and
  Kanoulas}]{leivaditi_benchmark_2020}
Spyretta Leivaditi, Julien Rossi, and Evangelos Kanoulas. 2020.
\newblock \href {http://arxiv.org/abs/2010.10386} {A {Benchmark} for {Lease}
  {Contract} {Review}}.
\newblock \emph{arXiv}.

\bibitem[{Lippi et~al.(2019)Lippi, Pałka, Contissa, Lagioia, Micklitz, Sartor,
  and Torroni}]{lippi_claudette_2019}
Marco Lippi, Przemysław Pałka, Giuseppe Contissa, Francesca Lagioia,
  Hans-Wolfgang Micklitz, Giovanni Sartor, and Paolo Torroni. 2019.
\newblock \href {https://doi.org/10.1007/s10506-019-09243-2} {{CLAUDETTE}: an
  automated detector of potentially unfair clauses in online terms of service}.
\newblock \emph{Artificial Intelligence and Law}, 27(2):117--139.

\bibitem[{Pedregosa et~al.(2011)Pedregosa, Varoquaux, Gramfort, Michel,
  Thirion, Grisel, Blondel, Prettenhofer, Weiss, Dubourg, Vanderplas, Passos,
  Cournapeau, Brucher, Perrot, and Duchesnay}]{pedregosa_2011}
F.~Pedregosa, G.~Varoquaux, A.~Gramfort, V.~Michel, B.~Thirion, O.~Grisel,
  M.~Blondel, P.~Prettenhofer, R.~Weiss, V.~Dubourg, J.~Vanderplas, A.~Passos,
  D.~Cournapeau, M.~Brucher, M.~Perrot, and E.~Duchesnay. 2011.
\newblock Scikit-learn: Machine learning in {P}ython.
\newblock \emph{Journal of Machine Learning Research}, 12:2825--2830.

\bibitem[{Qi et~al.(2020)Qi, Zhang, Zhang, Bolton, and
  Manning}]{qi-etal-2020-stanza}
Peng Qi, Yuhao Zhang, Yuhui Zhang, Jason Bolton, and Christopher~D. Manning.
  2020.
\newblock \href {https://doi.org/10.18653/v1/2020.acl-demos.14} {{S}tanza: A
  python natural language processing toolkit for many human languages}.
\newblock In \emph{Proceedings of the 58th Annual Meeting of the Association
  for Computational Linguistics}.

\bibitem[{Rabelo et~al.(2020)Rabelo, Kim, Goebel, Yoshioka, Kano, and
  Satoh}]{rabelo_coliee_nodate}
Juliano Rabelo, Mi-Young Kim, Randy Goebel, Masaharu Yoshioka, Yoshinobu Kano,
  and Ken Satoh. 2020.
\newblock \href
  {https://sites.ualberta.ca/~rabelo/COLIEE2021/COLIEE_2020_summary.pdf}
  {{COLIEE} 2020: {Methods} for {Legal} {Document} {Retrieval} and
  {Entailment}}.

\bibitem[{{The U.S. Securities and Exchange Commission}(2018)}]{sec_edgar_2018}
{The U.S. Securities and Exchange Commission}. 2018.
\newblock \emph{EDGAR® Public Dissemination Service Technical Specification}.

\bibitem[{Thorne et~al.(2018)Thorne, Vlachos, Christodoulopoulos, and
  Mittal}]{thorne_fever_2018}
James Thorne, Andreas Vlachos, Christos Christodoulopoulos, and Arpit Mittal.
  2018.
\newblock \href {https://doi.org/10.18653/v1/N18-1074} {{FEVER}: a
  {Large}-scale {Dataset} for {Fact} {Extraction} and {VERification}}.
\newblock In \emph{Proceedings of the 2018 {Conference} of the {North}
  {American} {Chapter} of the {Association} for {Computational} {Linguistics}:
  {Human} {Language} {Technologies}, {Volume} 1 ({Long} {Papers})}, pages
  809--819. Association for Computational Linguistics.

\bibitem[{Wu et~al.(2016)Wu, Schuster, Chen, Le, Norouzi, Macherey, Krikun,
  Cao, Gao, Macherey, Klingner, Shah, Johnson, Liu, Kaiser, Gouws, Kato, Kudo,
  Kazawa, Stevens, Kurian, Patil, Wang, Young, Smith, Riesa, Rudnick, Vinyals,
  Corrado, Hughes, and Dean}]{wu_googles_2016}
Yonghui Wu, Mike Schuster, Zhifeng Chen, Quoc~V. Le, Mohammad Norouzi, Wolfgang
  Macherey, Maxim Krikun, Yuan Cao, Qin Gao, Klaus Macherey, Jeff Klingner,
  Apurva Shah, Melvin Johnson, Xiaobing Liu, Łukasz Kaiser, Stephan Gouws,
  Yoshikiyo Kato, Taku Kudo, Hideto Kazawa, Keith Stevens, George Kurian,
  Nishant Patil, Wei Wang, Cliff Young, Jason Smith, Jason Riesa, Alex Rudnick,
  Oriol Vinyals, Greg Corrado, Macduff Hughes, and Jeffrey Dean. 2016.
\newblock \href {http://arxiv.org/abs/1609.08144} {Google's {Neural} {Machine}
  {Translation} {System}: {Bridging} the {Gap} between {Human} and {Machine}
  {Translation}}.
\newblock \emph{arXiv}.

\bibitem[{Zheng et~al.(2021)Zheng, Guha, Anderson, Henderson, and
  Ho}]{zheng_when_2021}
Lucia Zheng, Neel Guha, Brandon~R. Anderson, Peter Henderson, and Daniel~E. Ho.
  2021.
\newblock \href {https://arxiv.org/abs/2104.08671} {When {Does} {Pretraining}
  {Help}? {Assessing} {Self}-{Supervised} {Learning} for {Law} and the
  {CaseHOLD} {Dataset}}.
\newblock In \emph{Proceedings of the 18th {International} {Conference} on
  {Artificial} {Intelligence} and {Law}}. Association for Computing Machinery.

\end{thebibliography}
\bibliographystyle{acl_natbib}

\appendix

\clearpage

\section{Appendix}\label{sec:appendix}

\begin{table*}[t]
    \centering
    \fontsize{8pt}{7pt}\selectfont
    \setlength{\tabcolsep}{4pt}
    \renewcommand{\arraystretch}{1.4}
    \begin{tabularx}{\linewidth}{rlX}\toprule
        \# & Title & Hypothesis \\\midrule
        1 & Explicit identification & All Confidential Information shall be expressly identified by the Disclosing Party. \\
        2 & Non-inclusion of non-technical information & Confidential Information shall only include technical information. \\
        3 & Inclusion of verbally conveyed information & Confidential Information may include verbally conveyed information. \\
        4 & Limited use & Receiving Party shall not use any Confidential Information for any purpose other than the purposes stated in Agreement. \\
        5 & Sharing with employees & Receiving Party may share some Confidential Information with some of Receiving Party's employees. \\
        6 & Sharing with third-parties & Receiving Party may share some Confidential Information with some third-parties (including consultants, agents and professional advisors). \\
        7 & Notice on compelled disclosure & Receiving Party shall notify Disclosing Party in case Receiving Party is required by law, regulation or judicial process to disclose any Confidential Information. \\
        8 & Confidentiality of Agreement & Receiving Party shall not disclose the fact that Agreement was agreed or negotiated. \\
        9 & No reverse engineering & Receiving Party shall not reverse engineer any objects which embody Disclosing Party's Confidential Information. \\
        10 & Permissible development of similar information & Receiving Party may independently develop information similar to Confidential Information. \\
        11 & Permissible acquirement of similar information & Receiving Party may acquire information similar to Confidential Information from a third party. \\
        12 & No licensing & Agreement shall not grant Receiving Party any right to Confidential Information. \\
        13 & Return of confidential information & Receiving Party shall destroy or return some Confidential Information upon the termination of Agreement. \\
        14 & Permissible copy & Receiving Party may create a copy of some Confidential Information in some circumstances. \\
        15 & No solicitation & Receiving Party shall not solicit some of Disclosing Party's representatives. \\
        16 & Survival of obligations & Some obligations of Agreement may survive termination of Agreement. \\
        17 & Permissible post-agreement possession & Receiving Party may retain some Confidential Information even after the return or destruction of Confidential Information. \\
        \bottomrule
    \end{tabularx}
    \caption{List of hypotheses. The titles are only used for human readabilities.}\label{tab:hypotheses}
\end{table*}

\subsection{Details on Data Collection}\label{sec:appendix-data-collection}

In this section, we provide supplemental information regarding the data collection discussed in \cref{sec:system-data}.

As discussed in \cref{sec:system-formulation}, our dataset consists exclusively of non-disclosure agreements (NDAs) in order to incorporate more fine-grained hypotheses.
More specifically, we used unilateral or bilateral NDAs or confidentiality agreement between two parties.
We excluded employer-employee NDAs and those that are part of larger agreements (such as a confidentiality agreement inside a larger merger agreement), because they are quite different from the rest of NDAs.

We collected NDAs from Internet search engines and Electronic Data Gathering, Analysis, and Retrieval system (EDGAR).
For the collection from the search engines, we queried Google search engines with a search query ``\,``non-disclosure'' agreement filetype:pdf'' and downloaded the PDF files that the search engines returned.
We note that Google search engines in different domains return different results.
Therefore, we used seven domains from countries where English is widely spoken (US ``.com'', UK ``.co.uk'', Australia ``.com.au'', New Zealand ``.co.nz'', Singapore ``.com.sg'', Canada ``.ca'' and South Africa ``.co.za'').
Since collected PDFs contain irrelevant documents, we manually screened all 557 documents and removed all the irrelevant documents.
We also removed NDAs that do not have embedded texts (i.e., glyphs are embedded as an image) or those that have more than one columns, since they are difficult to preprocess.

For the collection from EDGAR, we first download all the filed documents from 1996 to 2020 in a form of daily archives\footnote{\url{https://www.sec.gov/Archives/edgar/Oldloads/}}.
We uncompressed each archive and deserialized files using regular expressions by referencing to the EDGAR specifications \cite{sec_edgar_2018}, which gave us 12,851,835 filings each of which contains multiple documents.
We then extracted NDA candidates by a rule-based filtering.
Using meta-data obtained during the deserialization, we extracted documents whose file type starts with ``EX'' (denotes that it is an exhibit), its file extension is one of ``.pdf'', ``.PDF'', ``.txt'', ``.TXT'', ``.html'', ``.HTML'', ``.htm'' or ``HTM'', and its content is matched by a regular expression ``(?<![a-zA-Z\.,\-"()]\textvisiblespace*)([Nn]on[-\textvisiblespace][Dd]isclosure)|(NON[-\textvisiblespace]DISCLOSURE)''.
We manually screened all 28,780 NDA candidates and obtained 236 NDAs.
All of the NDAs from EDGAR were either in HTML or plain text format.

\begin{figure}[tb]
    \centering
    \begin{subfigure}[b]{0.49\textwidth}
        \centering
        \includegraphics[width=\textwidth]{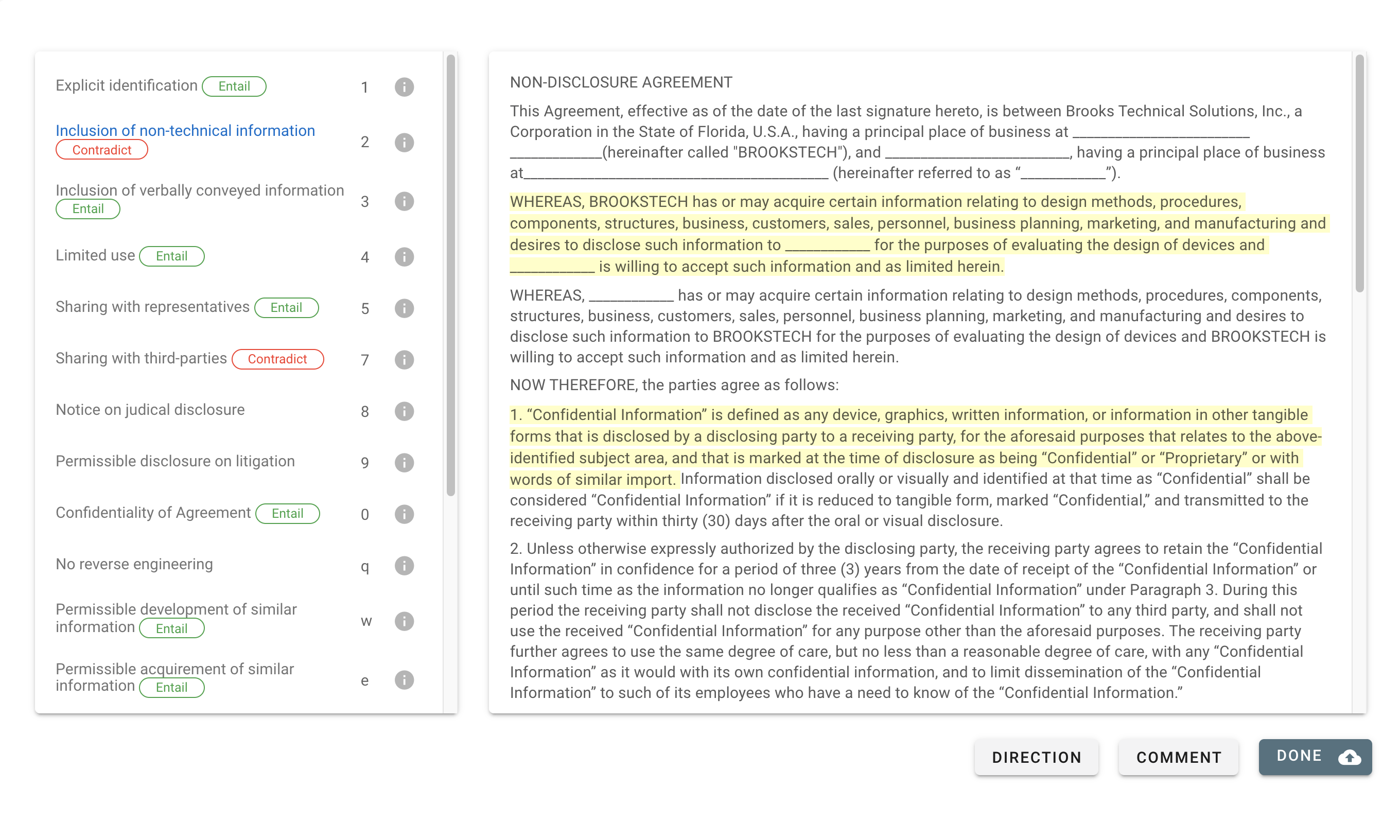}
        \caption{Evidence identification}
        \label{fig:annotation_ui-span}
    \end{subfigure}
    \begin{subfigure}[b]{0.49\textwidth}
        \centering
        \includegraphics[width=\textwidth]{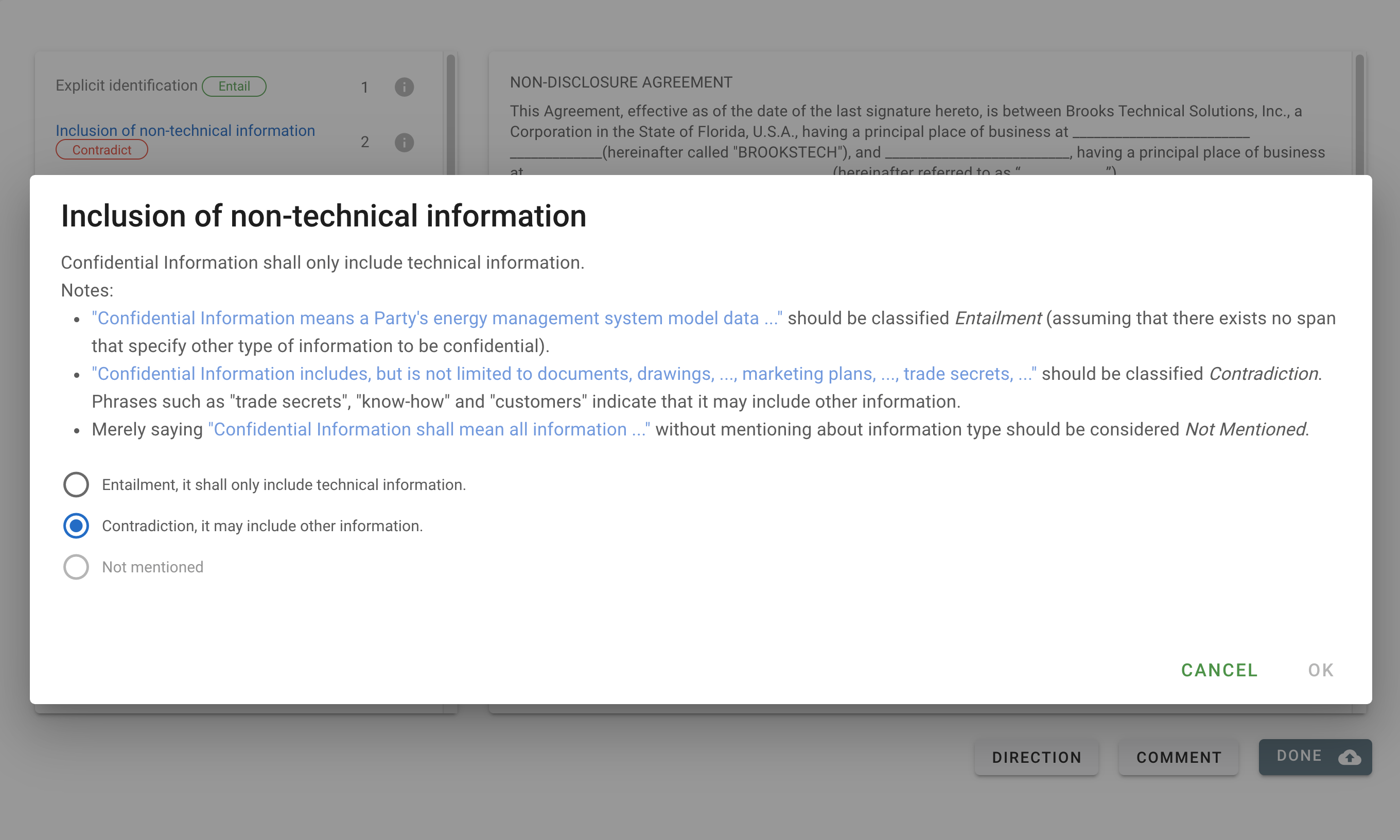}
        \caption{NLI}
        \label{fig:annotation_ui-nli}
    \end{subfigure}
    \caption{Question answering with evidence annotation interface}\label{fig:annotation_ui}
\end{figure}

\subsubsection{Details on Contract Annotation}

We developed 17 hypotheses by comparing different NDAs and had them reviewed by paralegals.
List of hypotheses can be found in \cref{tab:hypotheses}.

Since we employ a fixed set of hypotheses unlike existing NLI datasets, we employed an example-oriented annotation guideline for each hypothesis in order to improve annotation consistency.
Furthermore, we developed an annotation interface in order to efficiently and consistently annotate the NDAs.
The interface allows the users to select spans (\cref{fig:annotation_ui-span}) and then a NLI label (\cref{fig:annotation_ui-nli}).

Annotation was conducted by a computational linguistic researcher (the primary annotator) with a help from workers at Amazon Mechanical Turk.
We chose two workers at Amazon Mechanical Turk who were consistently performing well and asked them to redundantly annotate each document with a priority on coverage.
We merged annotated spans for each document.
Finally, the primary annotator reviewed the merged annotations and adjusted the annotations where necessary.
For the train split, the primary annotator only reviewed the annotated spans to judge NLI labels and to consolidate the span boundaries.
For most of the test split, the primary annotator went through the whole contracts to further improve coverage.
Most of the development dataset and some of the test dataset were annotated exclusively by the primary annotator without a help from the workers.
This allowed us to obtain consistent and high coverage annotations.

\subsection{Detailed Experiment Settings}

\subsubsection{Baselines}\label{sec:appendix-experiments-baseline}

We provide supplemental information of the baselines discussed in \cref{sec:baselines}.

For Doc TF-IDF+SVM, Span TF-IDF+Cosine and Span TF-IDF+SVM, we tokenized the input using Stanza \cite{qi-etal-2020-stanza} and extracted unigram TF-IDF vectors using Scikit-learn's \cite{pedregosa_2011} \texttt{TfidfVectorizer} with the default configuration (i.e., no stopwords apart from punctuations, minimum document frequencies of one, and smoothed inverse document frequencies).
For Doc TF-IDF+SVM and Span-TF-IDF+SVM, we used a Support Vector Machine \cite[SVM;][]{chang_libsvm_2011} with a linear kernel with the default hyperparameters implemented in Scikit-learn (i.e., $C=1.0$ with a stopping tolerance of $0.001$).

\begin{table*}[t]
    \centering
    \fontsize{8pt}{10pt}\selectfont
    \renewcommand{\arraystretch}{1.0}
    \begin{tabular}{cccc}\toprule
        Hyperparameter & BERT\textsubscript{\textit{base}} & BERT\textsubscript{\textit{large}} & DeBERTa\textsubscript{\textit{xlarge}} \\\midrule
        Batch size & 32 & 32 & 32  \\
        Learning rate & 1e-5, 2e-5, 3e-5, \textbf{5e-5}  & 1e-5, \textbf{2e-5}, 3e-5, 5e-5 & 5e-6, 8e-6, 9e-6, \textbf{1e-5}  \\
        AdamW's $\epsilon$ & 1e-8 & 1e-8 & 1e-6  \\
        Weight decay & 0.0, \textbf{0.1} & \textbf{0.0}, 0.1 & 0.01  \\
        Max. gradient norm & 1.0 & 1.0 & 1.0  \\
        Warmup steps & 0, \textbf{1000} & 0, \textbf{1000} & \textbf{50}, 100, 500, 1000  \\
        \# epochs & \textbf{3}, 4, 5 & \textbf{3}, 4, 5 & 3, \textbf{4}, 5  \\\midrule
        Min. \# surrounding tokens $n$ & \textbf{64}, 128 & 64, \textbf{128} & \textbf{64}, 128 \\
        Loss weight $\lambda$ & 0.05, 0.1, \textbf{0.2}, 0.4 & \textbf{0.05}, 0.1, 0.2, 0.4 & 0.05, 0.1, 0.2, \textbf{0.4} \\
        Use weighted NLI & \textbf{\texttt{True}}, \texttt{False} & \textbf{\texttt{True}}, \texttt{False} & \textbf{\texttt{True}}, \texttt{False} \\
        \bottomrule
    \end{tabular}
    \caption{Hyperparameter search space. The hyperparameters below the middle line are the hyperparameters specific to Span NLI BERT. The bold values denote the best hyperparameters in our experiment.}\label{tab:hyperparameters}
\end{table*}

For SQuAD BERT, we tried to be as faithful to a commonly used implementation as possible.
Thus, we implemented SQuAD BERT by implementing preprocessing and postprocessing scripts for the Huggingface's implementation\footnote{\url{https://github.com/huggingface/transformers/blob/0c9bae09340dd8c6fdf6aa2ea5637e956efe0f7c/examples/question-answering/run_squad.py}; We have slightly modified their implementation so that we have access to start/end token probabilities.}.
Because the SQuAD BERT only utilizes the first span even if a training example included multiple spans, we created an example for each span of each document-hypothesis pair.
Within the Huggingface's implementation, each example is further split into contexts with a fixed window size.
It is trained to point at starting and ending tokens of the span, or at \texttt{[CLS]} token when a span is not present.
Instead of allowing it to predict spans at arbitrary boundaries, we calculate a score for each of predefined spans by averaging token scores associated with the start and end of the span over different context windows.
This makes sure that its performance is not discounted for getting span boundaries wrong.

\subsubsection{Hyperparameters}\label{sec:appendix-experiments-hyperparameters}

For Span NLI BERT, we ran the same experiment ten times with different hyperparameters (\cref{tab:hyperparameters}).
Hyperparameter search spaces for BERT and DeBERTa have been adopted from \cite{devlin_bert_2019} and \cite{he_deberta_2021}, respectively.
For the SQuAD BERT baseline, we ran hyperparameter search over 18 hyperparameter sets as described in \cite{devlin_bert_2019}.

In both cases, we report the average score of three models with the best development scores.
Since NLI is more challenging than evidence identification, we used macro average NLI accuracy for the criterion.

The choice of weighted/unweighted NLI probablities was a part of our hyperparameters and we found that the best models (for BERT\textsubscript{\textit{base}}, BERT\textsubscript{\textit{large}} and DeBERTa\textsubscript{\textit{xlarge}}) preferred the weighted probablities.
The models with weighted probablities had on average 0.782 (BERT\textsubscript{\textit{base}}) and 0.803 (BERT\textsubscript{\textit{large}}) macro average NLI accuracies whereas the models with unweighted probablities had on average 0.458 (BERT\textsubscript{\textit{base}}) and 0.454 (\textsubscript{\textit{large}}) macro average NLI accuracies.
This implies that it is critical to incorporate the weighted probablities.

As for the loss weight $\lambda$, we found in pilot experiments that NLI starts to overfit faster than span detection, thus we searched values in $\lambda < 1$.
A possible hypothesis is that there is less diversity in teacher signal for NLI than that for evidence span detection; Contexts extracted from a single hypothesis-document pair have the same NLI label which could be somewhat redundant, whereas each context has a different span label.

\end{document}